\documentclass[letterpaper, 10 pt, journal, twoside]{IEEEtran}
\ifCLASSINFOpdf
\else
\fi

\usepackage{amsfonts}
\usepackage{amsmath,amssymb,amsfonts}
\usepackage{subcaption}
\usepackage{graphicx}
\usepackage{algorithm,algorithmic}
\usepackage{url}
\usepackage{booktabs}
\usepackage{comment}
\usepackage{todonotes}
\usepackage{multirow}
\usepackage{dblfloatfix}

\usepackage{pgfplots}
\pgfplotsset{compat=1.16}

\usepackage{xcolor}
\newcommand{\colortext}[2]{\textcolor{#1}{#2}}

\usepackage{lipsum} 

\usepackage[hidelinks]{hyperref} 

\begin{document}

\onecolumn
\thispagestyle{empty}
\begin{center}
    \null
    \vfill

    This paper has been accepted for publication in \textit{IEEE Robotics and Automation Letters}.

    This is the author's version of an article that has, or will be, published in this journal or conference.
    
    Changes were, or will be, made to this version by the publisher prior to publication.

    \vspace{2\baselineskip} 

    DOI: \href{https://doi.org/10.1109/LRA.2023.3336245}{10.1109/LRA.2023.33362450}
    
    IEEE Xplore: \href{https://ieeexplore.ieee.org/document/10327777}{https://ieeexplore.ieee.org/document/10327777}

    \vspace{2\baselineskip} 

    \vfill
    \vspace{5\baselineskip} 
    \footnotesize ©2023 IEEE. Personal use of this material is permitted. Permission from IEEE must be obtained for all other uses, in any current or future media, including reprinting/republishing this material for advertising or promotional purposes, creating new collective works, for resale or redistribution to servers or lists, or reuse of any copyrighted component of this work in other works.
\end{center}

\newpage

\twocolumn

\setcounter{page}{1} 

%
\title{Visual-Policy Learning through Multi-Camera View to Single-Camera View Knowledge Distillation for Robot Manipulation Tasks}
%
%
%

\author{Cihan Acar$^{1}$, Kuluhan Binici$^{2}$, Alp Tekirdağ$^{3}$  and Yan Wu$^{1}$%
\thanks{Manuscript received: July, 14, 2023; Revised September, 29, 2023; Accepted November, 5, 2023.
This paper was recommended for publication by Editor A. Faust upon evaluation of the Associate Editor and Reviewers' comments.
This work was supported by grant no. A19E4a0101 from the Singapore Government’s Research, Innovation and Enterprise 2020 plan (Advanced Manufacturing and Engineering domain) and administered by the Agency for Science, Technology and Research.
This work was also supported by the A*STAR Computational Resource Centre through the use of its high performance computing facilities} 
\thanks{$^{1}$C. Acar, Y. Wu  are with Institute for Infocomm Research (I2R), A*STAR, Singapore 138632
        {\tt\footnotesize acar\_cihan@i2r.a-star.edu.sg, wuy@i2r.a-star.edu.sg }}%
\thanks{$^{2} $K. Binici is with the School of Computing, National University of Singapore, Singapore
        {\tt\footnotesize kuluhan@u.nus.edu}}%
\thanks{$^{3} $A. Tekirdağ is is with the School of Computer Science and Engineering, Nanyang Technological University, Singapore
        {\tt\footnotesize alp001@e.ntu.edu.sg}}%

\thanks{Digital Object Identifier (DOI): {10.1109/LRA.2023.33362450}}
}

%
%

\markboth{IEEE Robotics and Automation Letters. Preprint Version. Accepted November, 2023}
{Acar \MakeLowercase{\textit{et al.}}: Visual-Policy Learning through Multi-Camera View to Single-Camera View Knowledge Distillation} 

%



\maketitle



\begin{abstract}
The use of multi-camera views simultaneously has been shown to improve the generalization capabilities and performance of visual policies. 
However, using multiple cameras in real-world scenarios can be challenging.
In this study, we present a novel approach to enhance the generalization performance of vision-based Reinforcement Learning (RL) algorithms for robotic manipulation tasks. 
Our proposed method involves utilizing a technique known as knowledge distillation, in which a ``teacher'' policy, pre-trained with multiple camera viewpoints, guides a ``student'' policy in learning from a single camera viewpoint. 
To enhance the student policy's robustness against camera location perturbations, it is trained using data augmentation and extreme viewpoint changes.
As a result, the student policy learns robust visual features that allow it to locate the object of interest accurately and consistently, regardless of the camera viewpoint.
The efficacy and efficiency of the proposed method were evaluated in both simulation and real-world environments. 
The results demonstrate that the single-view visual student policy can successfully learn to grasp and lift a challenging object, which was not possible with a single-view policy alone.
Furthermore, the student policy demonstrates zero-shot transfer capability, where it can successfully grasp and lift objects in real-world scenarios for unseen visual configurations [Video attachment: \url{https://youtu.be/CnDQK9ly5eg}].
\end{abstract}


%
\IEEEpeerreviewmaketitle

\section{Introduction}
%
%
%
%

\IEEEPARstart{I}{n}  recent years, the field of deep reinforcement learning (DRL) has seen rapid progress in robot control and manipulation applications, with vision-based RL algorithms being particularly noteworthy due to their ability to learn directly from high-dimensional visual inputs such as images, point clouds or videos~\cite{levine2016end}-\nocite{kalashnikov2018qt}\cite{levine2018learning}. 
However, one of the most significant challenges for these algorithms remains the generalization of learned skills to new and diverse situations. 
To address this issue, researchers have employed various data randomization~\cite{tobin2017domain} and augmentation methods, including changes to the background color, texture, and other parameters~\cite{yarats2020image}\nocite{laskin2020reinforcement}-\nocite{hansen2021generalization}\cite{srinivas2020curl}. 
While these techniques have shown effectiveness in learning representations of high-dimensional data, their generalization capabilities are limited to the features acquired from a fixed viewpoint perspective only.
\begin{figure}[t]
    \centering
    \includegraphics[scale=0.07]{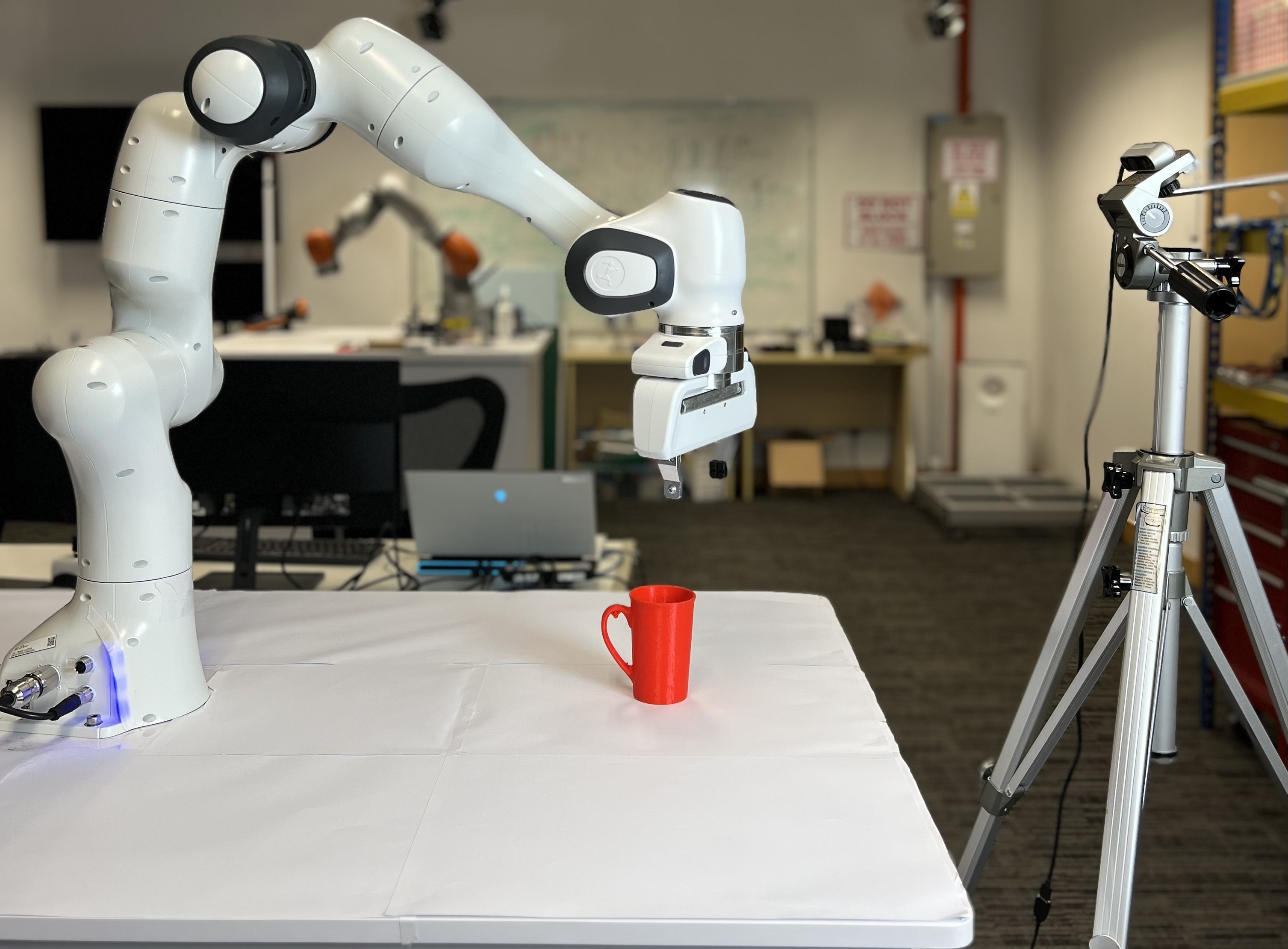}
    \caption{\small An illustration of a single-camera visual policy setup for robot manipulation tasks, with a front-facing third-person camera.}
    \label{fig:setup}
    \vspace{-15pt}
\end{figure}

To overcome this limitation, researchers have explored various techniques to enhance the generalization capabilities of vision-based learning algorithms~\cite{sadeghi2017sim2real}-\nocite{stadie2017third}\cite{shang2021self}. 
One approach involves training the algorithm using multiple viewpoints to recognize objects from different perspectives. 
For example, to develop visuomotor manipulation skills, researchers often employ a third-person camera view that captures the robot and environment from the outside, as well as a first-person (egocentric) camera that is linked to the robot's end-effector. The third-person camera offers a broad, fixed perspective of the environment and the robot, leading to a more comprehensive understanding. On the other hand, the egocentric camera provides a closer but limited observation since it cannot fully encompass the robot arm and the environment, potentially causing vital visual features to be obscured.
However, by concurrently utilizing both egocentric and third-person cameras, these limitations can be overcome, resulting in enhanced generalization capabilities and improved visual policy performance~\cite{hsu2022vision}\cite{jangir2022look}.
In a simulation environment, it is possible to use multiple third-person and egocentric cameras together to obtain better results without having any limitations. 
In contrast, deploying multiple camera devices in real-world scenarios presents significant challenges, including the need for additional hardware components and design constraints.  
In addition, coordinating and synchronizing multiple cameras to capture frames simultaneously can be technically complex and time-consuming. 

In this paper, we propose a method for learning a robust manipulation policy that can generalize to unseen visual configurations, using a single camera.  
Our approach leverages knowledge distillation to transfer privileged multiple-view information from a teacher policy to a student policy that observes the task using information only from a single camera. 
By learning from the teacher policy's rich multi-view information, the student policy can acquire robust features that can consistently locate the object of interest, regardless of the single-camera viewpoint.
First, a teacher policy is trained with multiple views gathered from fixed viewpoints to learn more generalizable representations. 
Then, a student with a single-camera viewpoint is trained to imitate the action of the teacher policy. 
To enhance robustness against camera location, the camera locations and parameters are constantly changed, in addition to data augmentation techniques such as random cropping and color jiggering. 
Furthermore, we employ a straightforward curriculum training approach for continuous policy learning by gradually expanding the student's camera parameter range.
Notably, camera randomization alone is insufficient on its own to ensure robust learning. 
Our experiments demonstrate that substantial camera randomization, without the guidance of knowledge distillation, fails to enable policies to acquire the essential knowledge and skills required for successful task performance.
Consequently, the single-camera student can learn the same actions as the teacher from heavily augmented and perturbed camera positions. 
This enables the student policy to perform manipulation tasks using a single-camera observation and as a result, makes it more practical and efficient for real-world applications.
To this end, the main contributions can be summarized as follows:
\begin{itemize}
    \item A novel and robust visual-policy learning by transferring privileged multiple-view information of the teacher to the single-camera view student.
    \item The use of randomized camera views during training, allowing the student policy to learn to adapt to different camera angles and positions.
    \item Investigation on the effect of the number of third-person camera views for knowledge distillation from the multiple-view teacher to the single-view student.
\end{itemize}

\section{RELATED WORK}
\label{sec:Background}
\subsection{Vision-based Reinforcement Learning}
The overall goal of vision-based reinforcement learning is to enable agents to learn from visual inputs, such as images or videos, in order to perform tasks in complex environments. 
To extract high-level features and improve the sample efficiency of vision-based RL algorithms, CURL method is proposed~\cite{yarats2020image}. This method employs contrastive loss to ensure consistency between an image and its augmented version. Many works also focus on data augmentation methods such as random shift, random crop, color jitter, applying Gaussian noise, and random convolutions for learning robust representations from the raw input image~\cite{laskin2020reinforcement}-\nocite{hansen2021generalization}\cite{srinivas2020curl}.
Data augmentation and CURL are also successfully implemented for real-world robotic tasks~\cite{zhan2020framework}.
One limitation of these methods is that they may not be adequate to handle more intricate forms of variation, such as variations in viewpoint or camera parameters.
\vspace{-10pt}
\subsection{Knowledge Distillation}
The procedure of using the knowledge of a pre-trained neural network model to enhance the capabilities of another one is referred to as knowledge distillation (KD)~\cite{hinton2015distilling}. 
Most commonly the pre-trained model, called the ``teacher", is used to provide additional annotation for the data samples to be used along with the ground truth labels while training another model from scratch.  
This enriches the limited information conveyed by the ground truth labels. As a result, the model trained with such enhanced information called the ``student", can achieve performance beyond its original capacity and even outperform the teacher.

In the context of reinforcement learning, KD has been employed for various tasks, such as policy distillation~\cite{rusu2015policy} which proposes distilling the knowledge of a large and accurate policy network, called the teacher, into a smaller and faster policy network, called the student.
Additionally, it is worth noting that similar concepts have been extended to scenarios where a single student agent can leverage multiple expert teachers, each specializing in individual tasks, allowing for more diverse and versatile knowledge transfer and multi-task learning~\cite{schmitt2018kickstarting}\cite{haarnoja2023learning}.
\vspace{-10pt}
\subsection{Privileged Provision}
Recently, KD is utilized in learning vision-based urban driving~\cite{chen2020learning}\cite{zhang2021end}, dexterous hand manipulation~\cite{jain2019learning}, in-hand object reorientation~\cite{chen2022system}, stacking diverse objects~\cite{lee2021beyond} and quadrupedal locomotion over challenging environments~\cite{lee2020learning}\cite{miki2022learning} where the teacher policy that has access to privileged information in a simulation environment is used to guide the student policy that has only access to sensory information available in the real world.
The student policy is then deployed and successfully tested in a variety of challenging real-world environments.
Another approach~\cite{Sorokin2022}\cite{agarwal2023legged} is to train the teacher in a simplified virtual environment with basic geometry and rendering, enabling faster learning. 
Then, a student is trained in a high-fidelity simulation featuring realistic sensors with the guidance of the teacher before being deployed in the real world.
Similarly, KD is also applied for zero-shot generalization of visual policies~\cite{fan2021secant}.
At first, weak image augmentation is used to train an expert policy.
Then, using imitation learning with strong augmentations, a student network learns to mimic the expert policy.
Even though it has been demonstrated that strong augmentations can offer robust representation learning for a range of visual policy tasks in diverse simulation contexts, generalization capabilities are restricted to features acquired through a single fixed viewpoint.
\begin{figure*}[ht]
    \centering
    \includegraphics[scale=0.70]{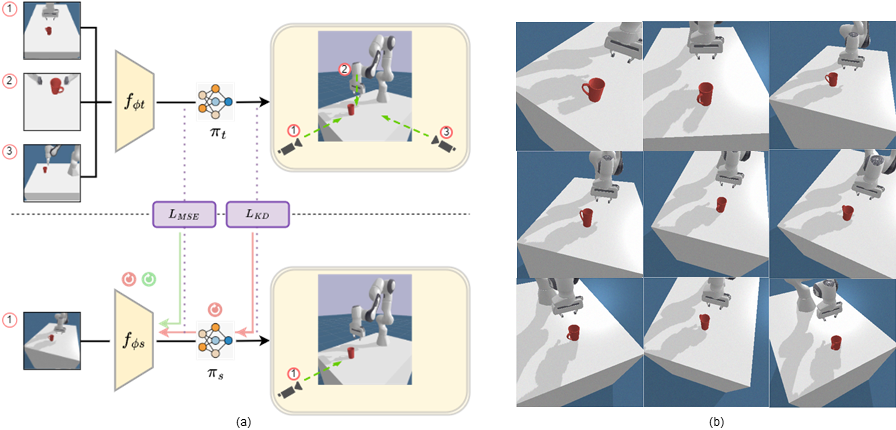}
    \caption{\small The figure depicts the overview of our framework for training the teacher  $\pi_{t}$ and student $\pi_{s}$ policies in part (a) and an example of randomized third-person camera views used to train the student policies in the visual-policy learning task in part (b). In part (a), we illustrate the flow of information and learning between the teacher and student policies. The teacher policy is trained using a  delayed reward scheme and is used to guide the training of the student policy. In part (b), we show a collection of sampled third-person camera views used to train the student policy, which are randomized by camera position and parameters to introduce variability in the training data and improve the generalization capability of the student policy. }
    \label{fig:frame_work}
    \vspace{-10pt}
\end{figure*}
\section{PRELIMINARIES}
The reinforcement learning environment is defined as a Markov Decision Process (MDP) tuple $\langle S,A,p,r,\gamma \rangle$: $S$ is the continuous state space, $A$ is continuous action space, $p(s_{t+1}|s_t,a_t)$ is the state transition model where probability density of the next state  $s_{t+1} \in S$ given $s_t \in S$ and $a_t \in A$, $r$ is the reward function, and $\gamma \in (0, 1]$ is the discount factor. 
Episode length is denoted by $T$.
The objective of an RL algorithm is to find an optimal policy $\pi: S \rightarrow A$ to maximize the expected return. 
  
\par
In order to determine the policy distribution $\pi(a_t|s_t)$, we employ an actor-critic model.
As our algorithm of choice, we use the Soft Actor-Critic (SAC)~\cite{haarnoja2018soft}, a state-of-the-art off-policy maximum entropy deep reinforcement learning algorithm due to its good performance in sparse reward setups.
The optimal policy equation of SAC is shown below:
\begin{equation}
\label{eq:SAC}
\pi^* = \arg\max_{\pi} \sum_{t} \mathbb{E}_{(s_t, a_t) \sim \rho_{\pi}} \left[ r(s_t, a_t) + \alpha H(\pi(\cdot | s_t)) \right]
\end{equation}
where the temperature parameter, denoted as $\alpha$, controls the stochasticity of the optimal policy by determining the relative importance of the entropy term compared to the reward.
\section{APPROACH}
\subsection{Problem Formulation}
In visual policy learning, the agent perceives the environment through cameras, and therefore, the environment is only partially observable. 
This makes the problem formulation a Partial Observability Markov Decision Process (POMDP) instead of a standard MDP. 
As a result, the state vectors $s_t$ in (\ref{eq:SAC}) are replaced with observations, denoted as $o$. 
These observations are obtained by encoding RGB images collected from the environment $[I^{1},...I^{N}]$ via a function $f$ before being provided to the  policy module:
\begin{equation}
    o := f([I^{1},...I^{N}]) \quad  \text{where} \quad  I\in \mathbb{R}^{C \times H \times W}
\end{equation}
Typically, $f$ is parameterized by a neural network (NN) $\phi$ and learned using gradient-based optimization techniques. 
Since neural networks generalize poorly when training data lacks diversity, increasing the number of unique views $N$ by using more cameras can improve the performance. 
As a result, there is a natural trade-off between the hardware complexity of the robotics system and performance since utilizing multiple camera devices to process diverse views of the environment typically provides robust policies against domain variations at the expense of increased hardware demand and design complexity. 
To break such a trade-off, we propose training a privileged teacher policy $\pi_{t}(a|o_t)$ using multiple camera views from the simulation environment and later distilling the learned policy to train a non-privileged student policy $\pi_{s}(a|o_s)$ that works with only single-view observations. 
In this way, while the hardware requirement of deploying $\pi_{s}$ is limited to a single camera, it can provide similar performance to $\pi_{t}$. 

Our objective is to find the optimal student policy $\pi_{s}^{*}$ that minimizes the knowledge distillation (KD) loss. 
The KD loss measures the discrepancy between the student policy and the teacher policy:
\begin{equation}
\pi_{s}^{*} := \underset{\pi_{s}}{argmin}\,\,L_{KD}(\pi_s,\pi_t)
\end{equation}
\vspace{-20pt}
\subsection{Implementation Details} \label{implementation_details}
The teacher operates with images collected from a single egocentric camera ($I_{e}$) and $N$ stationary third-person view cameras ($[I_{t}^{\theta_1},...I_{t}^{\theta_N}]$) while the student uses the view of a single third-person view camera ($I_{t}^{\theta_k}$).
\begin{equation}
\begin{split}
    o_{t} &:= f_{\phi t}(I_{e}, [I_{t}^{\theta_1},...I_{t}^{\theta_N}]) \\
    o_{s} &:= f_{\phi s}(I_{t}^{\theta_k}) \quad \textrm{where} \quad \theta_k \sim U(0,\theta_{max})
\end{split}
\end{equation} To make the student robust against environmental perturbations, we apply domain randomization and sample the camera angle $\theta_k$ randomly from a uniform distribution that ranges from $0$ to $\theta_{max}$. 
Both the teacher and the student have the same network architecture comprised of a 12-layer convolutional encoder with a single-layer self-attention module and a SAC policy.
We use PyTorch~\cite{paszke2019pytorch} implementation of SAC as~\cite{zhan2020framework}.
\par\noindent
\textbf{Knowledge Distillation:} 
We apply the principles of DAgger~\cite{latexross2011reduction}, which reduces distributional shifts in learned policies by incorporating data from an expert policy, to distill the knowledge of the teacher policy into the student policy.
Different than DAgger, our expert is not a human, instead we use a pre-trained teacher policy.
To train the student, we use two sources of information: the actions and encoded features provided by the teacher. 
To distill the first information into the student, the mean squared error (MSE) between the student $\pi_{s}$ and teacher $\pi_{t}$ policies is minimized: 
\begin{equation}
L_{KD} :=\left\|\pi_{s}(o_s) - \pi_{t}(o_t)\right\|
\label{eq:KD}
\end{equation}
The student policy $\pi_{s}$ is updated by minimizing this loss using the transition batch sampled from experiences replay buffer \( \mathcal{D} \).
The updated student policy is then used to collect more data, which is added to \( \mathcal{D} \) and used for the next iteration of the policy update.
During the training of the teacher, only the loss of the critic is used to update the weights of the encoder network for learning the representation space.   
For training the student encoder, instead of using the gradients of critic to update the student encoder, we implement two methods.
The first one is inspired by~\cite{tung2019similarity} where we align the pairwise similarity matrix of teacher observations $C_{t}$  with that of the student observations $C_{s}$ by minimizing the following loss:
\begin{equation} \label{eq:pairwise}
    L_{sim} := \sum_{i}\sum_{j}\|C_{T}^{ij} - C_{S}^{ij}\|_{F}^{2}
\end{equation}
Different to \cite{tung2019similarity}, We use the cosine distance to quantify the pairwise similarities between different observations:
\begin{equation}
    C := \frac{\overline{C}}{\|o_{[i,:]}\|\|o_{[i,:]}\|^{T}}; \quad \overline{C} := oo^{T}; \quad o \in \mathbb{R}^{b\times(z)}
\end{equation}
Here $o$ is the flattened output tensor of the feature extractor $f_\phi$. 
In the second one, the distillation objective is to minimize the Euclidean distance between the features of the teacher and student. 
Thus, the second method tries to minimize the sum of all the squared differences between the teacher encoder output and the predicted student output:
\begin{equation} \label{eq:MSE}
L_{MSE} := \|o_t - o_s\|_{2}
\end{equation}
Our results, to be presented in the next section, show that minimizing the Euclidean distance between features yields superior performance for the student policy when compared to the pairwise similarity method. 
This finding suggests that minimizing the Euclidean distance provides better distillation for learning high-level visual features.
\begin{algorithm}[!t]
\caption{Multi-View to Single-View KD}
\label{alg:your_algorithm}
\begin{algorithmic}[1]
\small 
\STATE \textbf{Input:} multi-view teacher policy $\pi_t$, single-view student policy $\pi_s$, feature extractors $f_{\phi t}$ and $f_{\phi s}$, a buffer containing fixed expert demonstrations $\mathcal{B}_e$.
\STATE \#Train teacher policy $\pi_t$ and $f_{\phi t}$.
\FOR{$t$ in $1, \ldots, T_{\text{train}}$}
  \STATE Update $\pi_t$ and $f_{\phi t}$ based on  Eq. \ref{eq:SAC} using dataset $\mathcal{B}_e$.
  \STATE Execute policy $\pi_t$ and obtain new transitions $\mathcal{B}_t$.
  \STATE $\mathcal{B}_e \gets \mathcal{B}_e \cup \mathcal{B}_t$.
\ENDFOR
\vspace{0pt} 
\STATE \#Train student policy $\pi_s$ and $f_{\phi s}$ via KD.
\STATE Use $\pi_t$ to collect multi-camera view transitions $\mathcal{B}$.
\FOR{$t$ in $1, \ldots, T_{\text{distill}}$}
  \STATE Update $\pi_s$ and $f_{\phi s}$ to minimize the loss in Eq. \ref{eq:KD} and Eq. \ref{eq:MSE} using observations in $\mathcal{B}$.
  \STATE Execute policy $\pi_s$ and obtain new transitions $\mathcal{B}_s$.
  \STATE $\mathcal{B} \gets \mathcal{B} \cup \mathcal{B}_s$.
    \IF{Episode is done}
    \STATE Randomize camera viewpoint.
  \ENDIF
\ENDFOR
\vspace{-1pt} 
\end{algorithmic}
\end{algorithm}
\par\noindent
\textbf{Curriculum Learning:} To make the student robust against environmental perturbations we apply domain randomization and permute the camera angle $\theta_k$ and camera parameters.
We utilize a simple curriculum training that gradually increases the range of camera parameters of the student. 
Initially, when training begins, the camera parameter range closely aligns with that of the teacher.
As training progresses, we incrementally widen the parameter range every 50 episodes (up to the maximum parameter range) and sample camera parameters randomly from this expanded distribution. 
Our intuition is that gradually increasing the camera parameters increases the difficulty level of the task.
It is also possible to utilize more efficient automatic curriculum learning methods such as ALP-GMM~\cite{portelas2020teacher}, Automatic Curriculum Learning through
Value Disagreement~\cite{zhang2020automatic}, where the complexity of the task is determined by the performance of student policy.
\par\noindent
\textbf{Data Augmentation:} To improve the generalization ability of the policies, we employ two techniques for data augmentation during training: pixel shifting and color augmentation. 
These techniques help to expose the models to a more diverse range of visual inputs, which in turn can improve their ability to perform the task in a wider variety of scenarios. 
Specifically, we randomly shift the pixels of the images in two dimensions and apply color augmentation techniques such as changing the brightness, contrast, and saturation. 
By applying these techniques, we aim to make the policies more robust to changes in lighting, and other environmental factors that may be present in real-world scenarios. 
\par\noindent
\textbf{Setup Details:}  
Multiple cameras can be employed to capture various viewpoints and enhance the generalization capability of visual-policy learning. 
Our study focuses on the utilization of two-camera and three-camera setups to train an expert visual policy. 
The two-camera setup comprises a first-person (egocentric) camera attached to the robot's end-effector and a third-person camera positioned in front of the robot to capture a bird's eye view of the table and robot. 
The three-camera setup consists of a first-person camera and two third-person cameras positioned in front and on the side of the robot, respectively.
These cameras are positioned to offer orthogonal viewpoints. 
This arrangement minimizes information redundancy across cameras, maximizing the advantages of multi-view input.
During the training of the student policy, we employ only one of the third-person cameras to capture the visual information. 
More precisely, we use a single third-person camera placed in front of the robot but the camera parameters and view are randomly changed. 
By constantly changing the camera view and parameters, the training data becomes more diverse, and the student policy can learn to adapt to different camera angles and positions. 
This allows the student policy to generalize better to new, unseen camera views, which is crucial for applications where the camera position and orientation may not be known or fixed. 
\begin{figure*}[ht]
    \captionsetup{font=small}
    \centering
    \includegraphics[scale=0.41]{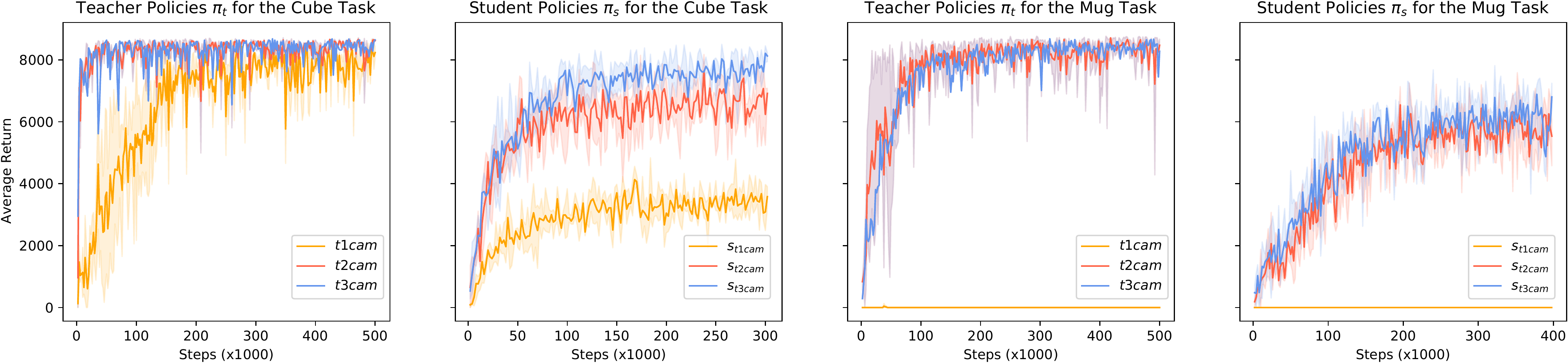}
    \caption{ Training results of teachers and distilled student policies for lifting cube and mug objects in 3 separate runs, each using different random seeds. The solid curves represent the mean returns, while the shaded regions indicate the range between the minimum and maximum returns. $t_{1cam}$, $t_{2cam}$, and $t_{3cam}$ refer to the single-camera view, two-camera view, and three-camera view teacher policies, respectively. $s_{t1cam}$, $s_{t2cam}$, and $s_{t3cam}$ refer to the student policies distilled from $t_{1cam}$, $t_{2cam}$, and $t_{3cam}$ teacher policies, respectively. }
    \label{fig:simulation_results}
    \vspace{-15pt}
\end{figure*}
\begin{table*}[!b]
    \caption{\small Average reward and success rates of cube and mug lifting task across 100 random view trials for 3 different runs \colortext{black}{per task each using different random seeds.} 
    In the fixed-angle setup, no perturbations were made to the front camera position. However, for the random angle settings, random front-camera views were introduced. 
      $\pi_{t1cam}$, $\pi_{t2cam}$, and $\pi_{t3cam}$ are the single, two, and three-camera view teacher policies, respectively.
       $\pi_{s_{t1cam}}$, $\pi_{s_{t2cam}}$, and $\pi_{s_{t3cam}}$ denotes the student policies distilled from the corresponding $\pi_{t1cam}$, $\pi_{t2cam}$, and $\pi_{t3cam}$ teacher policies, respectively.                            SCV: Single-Camera View, MCV: Multi-Camera View }
 \label{tab:simulation_evaluation}      
\resizebox{1\textwidth}{!}{
\begin{tabular}{clcccc}
\hline
\multirow{2}{*}{\textbf{Evaluation}} & \multicolumn{1}{c}{\multirow{2}{*}{\textbf{Methods}}} & \multicolumn{2}{c}{\textbf{Cube}}                                              & \multicolumn{2}{c}{\textbf{Mug}}                                               \\ \cline{3-6} 
                                     & \multicolumn{1}{c}{}                                  & \multicolumn{1}{l}{Average Reward (R)} & \multicolumn{1}{l}{Success Rate (\%)} & \multicolumn{1}{l}{Average Reward (R)} & \multicolumn{1}{l}{Success Rate (\%)} \\ \hline
\multirow{6}{*}{Fixed Angle}         & $\pi_{t1cam}$                                                 & 7954.0 $\pm$ 189.5                           & 96.5 $\pm$ 3.5                              & 0                         & 0.0                                   \\
                                     & $\pi_{t2cam}$                                                 & 8320.6 $\pm$  269.9                          & 98.6 $\pm$ 2.3                              & 8440 $\pm$ 111.4           & 99.3 $\pm$ 0.57                                      \\
                                     & $\pi_{2cam}$ (\cite{jangir2022look} cross-view attn.)         & 8397.6 $\pm$  142.5                          & \textbf{99.3 $\pm$  1.1}                             & \textbf{8574$\pm$ 83.8}            & 99.3 $\pm$  1.1                                   \\
                                     & $\pi_{t3cam}$                                                 & \textbf{8420.0 $\pm$  64.5}                  & 99.3 $\pm$ 5.7                     & 8543 $\pm$ 121.3         & \textbf{100 $\pm$  0.0}                                  \\
                                     & $\pi_{s_{t1cam}}$ (ours-SCV)                                  & 4285.6 $\pm$  321.3                          & 93.0 $\pm$ 3.6                              & –                         & –                                     \\
                                     & $\pi_{s_{t2cam}}$ (ours-MCV)                                  & 8015.6 $\pm$  231.1                          & 96.6 $\pm$ 0.5                              & 7431 $\pm$ 244.4          & 89.6 $\pm$  2.5                                   \\
                                     & $\pi_{s_{t3cam}}$ (ours-MCV)                                  & 8388.6 $\pm$  158.3                          & 99.0 $\pm$ 1.7                              & 7567 $\pm$ 112.8          & 91.6$\pm$  1.1                          \\ \hline
\multirow{6}{*}{Random Angle}        & $\pi_{t1cam}$                                                 & 112.0 $\pm$ 96.9                             & 02.0 $\pm$ 2.0                                               & 0                         & 0.0                                   \\
                                     & $\pi_{t2cam}$                                                 & 1597.6 $\pm$  352.5                   & 23.3$\pm$ 3.2                                              & 2216.3$\pm$ 1409.1        & 30.6  $\pm$ 18.8                                \\
                                     & $\pi_{2cam}$ (\cite{jangir2022look} cross-view attn.)          & 2821.6 $\pm$  2201.3                  & 39.6$\pm$ 29.1                                    & 2909.0$\pm$ 672.2         & 37.3 $\pm$ 9.1                                 \\
                                     & $\pi_{t3cam}$                                                 & 5419.0 $\pm$  916.7                    & 68.7$\pm$ 12.2                          & 2818.0 $\pm$ 171.6                          & 41.3  $\pm$  3.2                                  \\
                                     & $\pi_{s_{t1cam}}$ (ours-SCV)                                  & 3673.0 $\pm$  475.1                         & 84.6  $\pm$ 2.3                                 & –                                      & –                                     \\
                                     & $\pi_{s_{t2cam}}$ (ours-MCV)                                  & 7110.3 $\pm$  219.4                      & 88.7    $\pm$ 3.2                      &  5450.6 $\pm$ 491.7                  & 77.0 $\pm$  6.0                                 \\
                                     & $\pi_{s_{t3cam}}$ (ours-MCV)                                  & \textbf{7815.0 $\pm$149.2}                          & \textbf{96.0 $\pm$ 1.7  }           & \textbf{6334.3 $\pm$ 629.9}                          & \textbf{81.7 $\pm$  8.3}                         \\ \hline
\end{tabular}
}
\end{table*}
In our context, the visual policy is responsible for generating control actions that enable the robot to perform a given task. 
The action space includes control over the 3-DoF end-effector position along the X, Y, and Z axes, 1-DoF rotation control on the yaw-axis, and 1-DoF gripper control. 
We train both the teacher and student policies using 84x84 RGB camera images as observations. 
However, for tasks requiring finer details, it is possible to utilize higher-resolution observations during student policy training. 
This approach offers the advantage of reducing computational costs and accelerating training times when compared to training students from scratch using higher-resolution image observations.
\par\noindent
\textbf{Reward Scheme:} We train the teacher policies using a delayed reward scheme with a fixed time horizon, in which the agent gets rewarded only if the task is completed successfully.
Achieving a task successfully results in a +100 reward at each subsequent time step, and accumulates until the end of the episode, which encourages the agent to complete the task as quickly as possible.
\par\noindent
\textbf{Task Definition:} The tasks in our experiments require the robot to grasp either a cube or a mug and lift it to a predetermined height from its initial position. 
While cubes and spheres have a symmetrical shape that makes it easier for visual policies to learn grasping strategies, the lack of symmetry in the mug presents a greater challenge.
The task is further complicated by visual localization, which requires the agent to infer object positions solely from high-dimensional RGB camera pixel data. 
To increase the diversity of training data, objects are randomly spawned at different positions on the table for each episode. 
This ensures that the agent learns to adapt to different object positions, enhancing its ability to perform the task in novel scenarios.

\section{Results}
\label{sec:results}

\subsection{Simulation Experiments}
In order to evaluate the effectiveness of our proposed approach, we conducted experiments using a 7 degree-of-freedom Franka Panda Emika manipulator equipped with a two-fingered parallel gripper in a PyBullet simulation environment~\cite{coumans2013bullet}, based on code from~\cite{borja2022affordance}.
To train the teacher policies $\pi_{t}$, we utilized combinations of first-person, front, and side camera views, as depicted in Figure \ref{fig:frame_work}(a). 
The two-camera view teacher policy comprised a front camera labeled as (1) and an egocentric camera attached to the robot's end-effector labeled as (2) in Figure \ref{fig:frame_work}(a). 
Additionally, the three-camera view teacher policy included a side camera labeled as (3) in Figure \ref{fig:frame_work}(a) in addition to these.
Furthermore, we introduced a single-camera view teacher, which solely utilized a front camera. 
This single-camera view teacher provided a baseline for comparison and allowed us to evaluate the benefits of incorporating multiple camera views in the training process.

For each task, we used a 3D-mouse to collect 50 human demonstrations (total of 5K transitions), which were added to the replay buffer to aid in training the multi-camera view expert teachers.
It is important to note that these demonstrations were exclusively utilized for training the teacher policies and not for training the students.
This method of using demonstrations has been shown to be effective in improving the learning process and accelerating the convergence of the RL algorithms~\cite{zhan2020framework}.
During the training process, we keep only the most recent 50K transitions in the replay buffer.
The teacher policies were initially trained from fixed viewpoints. The student policies were subsequently distilled from these trained teacher policies.
We introduced a significant level of randomization to the camera position and parameters, such as field of view (fov) and aspect ratio to increase the robustness of the student policies $\pi_{s}$. 
PyBullet facilitated this by employing a synthetic camera specified by two 4 by 4 matrices: the view matrix and the projection matrix, allowing for the rendering of images from arbitrary camera positions for each episode. 
We provide a few examples in Figure \ref{fig:frame_work}(b) to showcase the different camera perspectives used during the training of the student policies in the simulation environment.
For all teachers and students trained, we used the same data argumentation methods described in section \ref{implementation_details}.
All simulation experiments are repeated 3 times, each utilizing a different random seed.
\par
The training results of teacher policies $\pi_{t}$ using different numbers of camera views and the corresponding results of the student policies distilled from them to lift the mug and cube objects are presented in Figure \ref{fig:simulation_results}.
For the cube object, all teacher policies have successfully learned the task, but the single-camera view teacher policy $\pi_{t1cam}$ had a lower performance than the multiple camera-view teacher policies $\pi_{t2cam}$ and $\pi_{t3cam}$. 
More importantly, for the mug object, the single-camera view teacher policy $\pi_{t1cam}$ was unable to effectively learn the task.
These results demonstrate the importance of utilizing multiple camera views for learning complex tasks such as lifting asymmetric objects effectively. 

Table \ref{tab:simulation_evaluation} presents the results of evaluating the performance of teacher and student policies on lifting the cube and mug objects across 100 fixed and random front-camera views. 
All results are averaged across three different runs.
The performances of seven different policies are compared, including single-camera teacher policy $\pi_{t1cam}$, two-camera multi-view teacher policy $\pi_{t2cam}$, three-camera multi-view teacher policy $\pi_{t3cam}$, and three student policies $\pi_{s_{t1cam}}$, $\pi_{s_{t2cam}}$, and $\pi_{s_{t3cam}}$, distilled from the corresponding teacher policies.
Additionally, we established a baseline by comparing with $\pi_{2cam}$, a cross-view attention-based two-camera view visual policy introduced by Jangir et al.~\cite{jangir2022look}.
The results indicate that for fixed and random front angles settings the students distilled from multi-view teachers, $\pi_{s_{t2cam}}$ and $\pi_{s_{t3cam}}$, outperform the students distilled from single-camera teachers, $\pi_{s_{t1cam}}$ in both cube and mug lifting tasks.
Furthermore, since the student policies are trained with extensive randomization of the camera positions and parameters, their performance surpasses that of the teacher policies and the cross-view attention policies in random camera view settings. However, it should be emphasized that without KD, applying only camera randomization resulted in a 0 success rate for the single-view agent in both tasks.
Notably, the multi-camera view to single-camera view distillation approach enables the single-camera view student policy to learn and perform the task of lifting the mug object, despite the single front-view camera teacher was unable to accomplish this. This highlights the effectiveness of distillation in enhancing the capabilities of the single-camera view policy.
In terms of teacher training time, multiple-view teacher policies require slightly more clock time due to increased data volume. 
The total training time for the students is the sum of distillation time and the time taken for training the teacher models. 
Consequently training $\pi_{s_{t3cam}}$ and $\pi_{s_{t2cam}}$ take 16\% and 9.1\% more time compared to $\pi_{s_{t1cam}}$, respectively.
\vspace{-3pt}
\par
\begin{figure}[ht]
    \centering
    \includegraphics[scale=0.35]{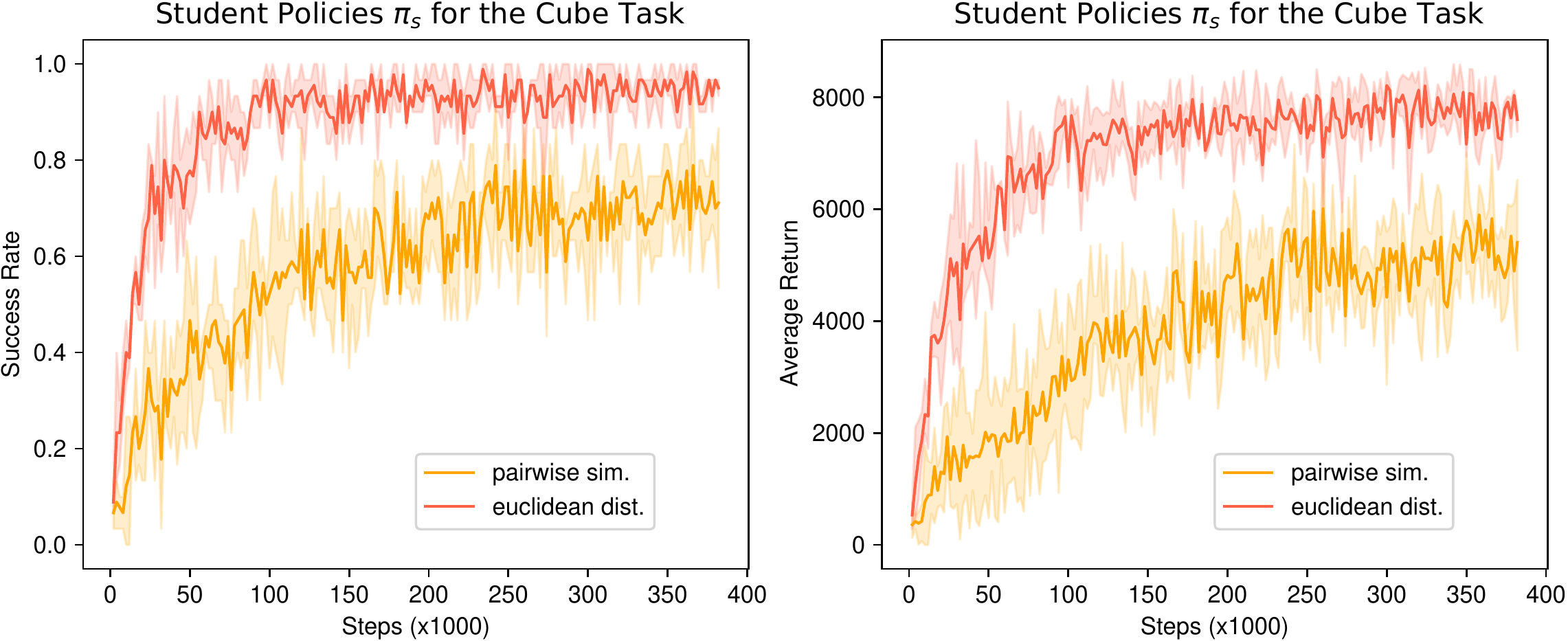}
    \caption{\small A comparison of two feature alignment methods utilized for knowledge distillation in our framework for lifting the cube object.} 
    \label{fig:feature_alligment}
    \vspace{-10pt}
\end{figure}

\noindent
\textbf{Ablation Studies:} In addition to our main experiments, we conducted ablation studies to investigate the effect of feature alignment between the teacher and student observations. 
As shown in Figure \ref{fig:feature_alligment}, we compared two different alignment methods: minimizing the euclidean distance between the features of the teacher and student (eq. \ref{eq:MSE}) and aligning the pairwise similarity matrix of teacher observations with that of the student observations (eq. \ref{eq:pairwise}). 
Our results indicate that the former method provides better results, suggesting that direct feature alignment is more effective for knowledge distillation in our framework. 
This analysis provides insights into the importance of feature alignment in knowledge distillation and its impact on the performance of the student policy.
\begin{figure*}[t]
    \centering
    \includegraphics[scale=0.5]{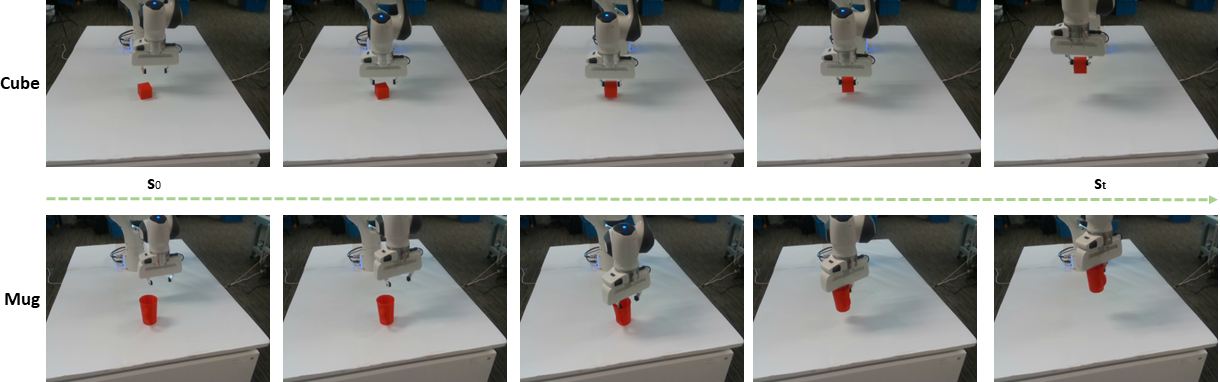}
    \caption{\small  Snapshots of the cube and mug lift task being performed by a robot in a real-world setup. The robot uses its single-view visual policy trained in
           simulation to successfully grasp and lift the objects. The robot's end-effector is located at the starting position, and the cube and mug objects are placed randomly on the table. }
    \label{fig:real_world_task}
    \vspace{-15pt}
\end{figure*}
\par\noindent

\subsection{Real Robot Experiments}
To validate the zero-shot policy transfer capability of the proposed framework, experiments were conducted in a real-world environment using the Panda robot with a Panda Hand gripper and an intel RealSense D435i RGBD camera, as shown in Figure \ref{fig:setup}.  
Polymetis robot control framework~\cite{lin2021polymetis} was used to control the robot and the gripper position. 
The visual policies trained for each of the aforementioned lifting tasks, i.e. mug and cube, in the simulation, were directly transferred to the real-world setting without any further training.  
Figure \ref{fig:real_world_task} shows snapshots of the robot successfully executing these tasks with a random camera viewpoint.

\begin{figure}[!b]
    \centering
    \begin{subfigure}{0.16\textwidth}
        \includegraphics[width=\textwidth]{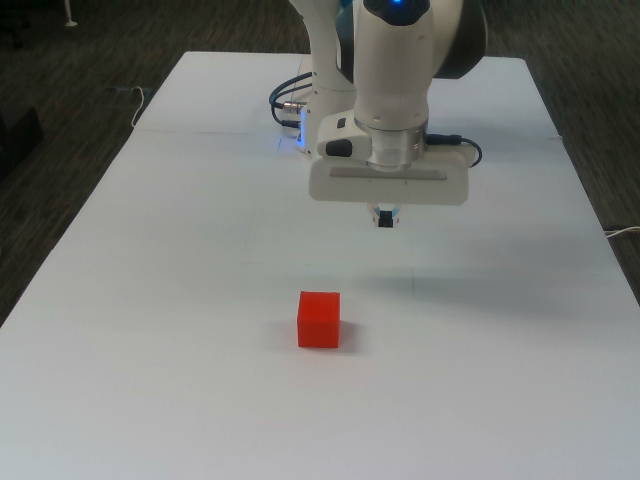} 
        \caption{ $\theta_{c1}$ }
        \label{fig:cube_real1}
    \end{subfigure}
    \begin{subfigure}{0.16\textwidth}
        \includegraphics[width=\textwidth]{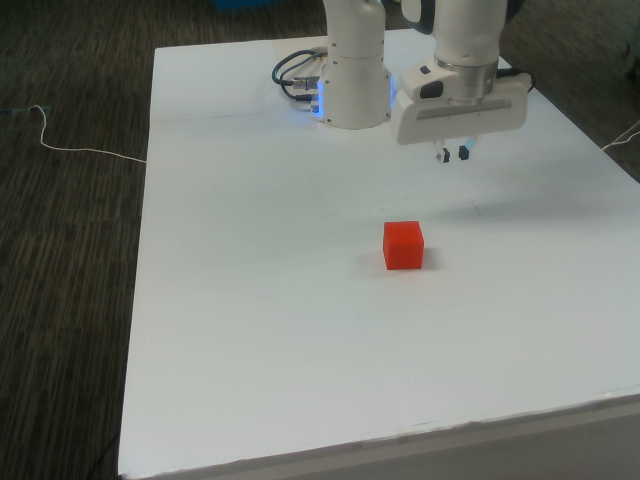}  
        \caption{$\theta_{c2}$}
        \label{fig:cube_real2}
    \end{subfigure}
    \begin{subfigure}{0.16\textwidth}
        \includegraphics[width=\textwidth]{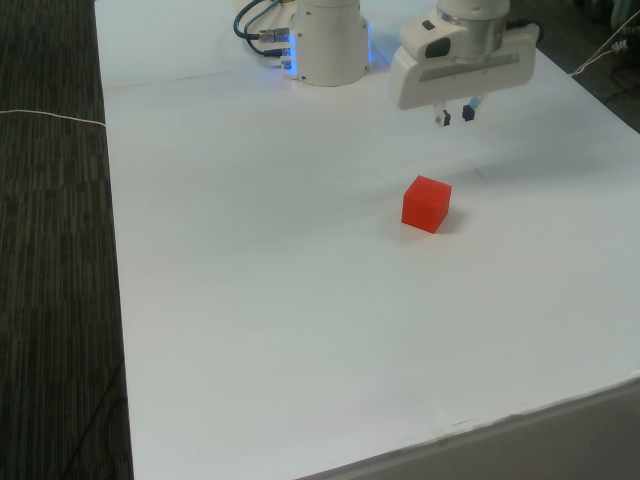} 
        \caption{$\theta_{c3}$}
        \label{fig:cube_real4}
    \end{subfigure}
    \begin{subfigure}{0.16\textwidth}
        \includegraphics[width=\textwidth]{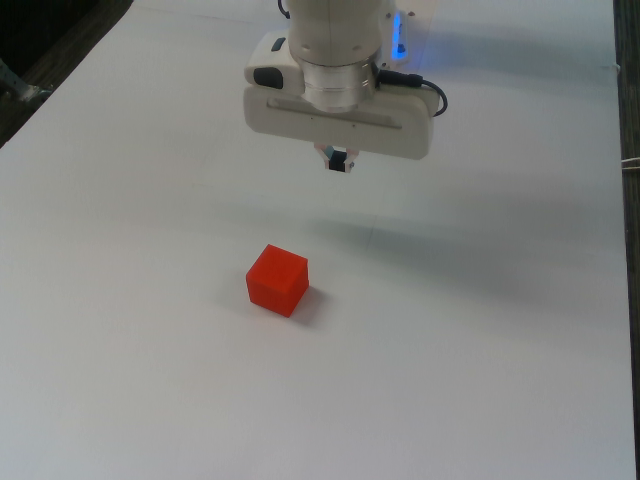}  
        \caption{$\theta_{c4}$}
        \label{fig:cube_real5}
    \end{subfigure}
    \caption{\small Cube lifting task executed with different random camera views in real-world experiment. The results for each view are given in Table \ref{tab:cube_success_rate_real}.}
    \hspace{4.4cm}
    \label{fig:cube_random_views}
\end{figure}
\begin{figure}[!b]
    \centering
    \begin{subfigure}{0.16\textwidth}
        \includegraphics[width=\textwidth]{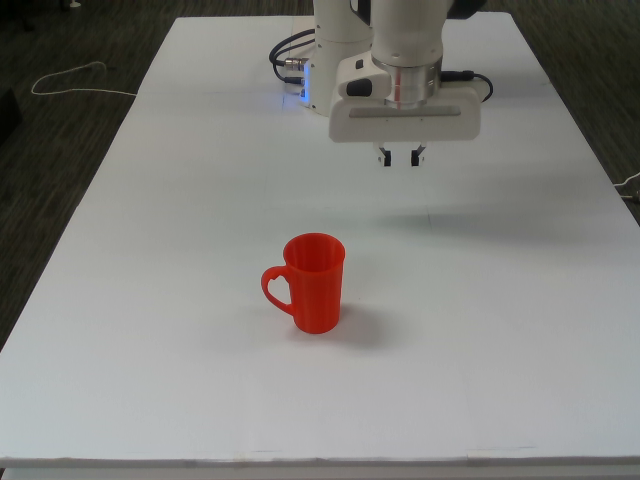} 
        \caption{$\theta_{m1}$}
        \label{fig:mug_real1}
    \end{subfigure}
    \begin{subfigure}{0.16\textwidth}
        \includegraphics[width=\textwidth]{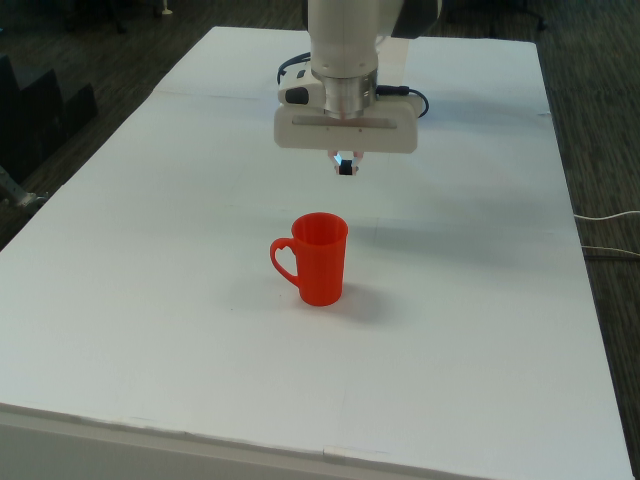}  
        \caption{$\theta_{m2}$}
        \label{fig:mug_real2}
    \end{subfigure}
    \begin{subfigure}{0.16\textwidth}
        \includegraphics[width=\textwidth]{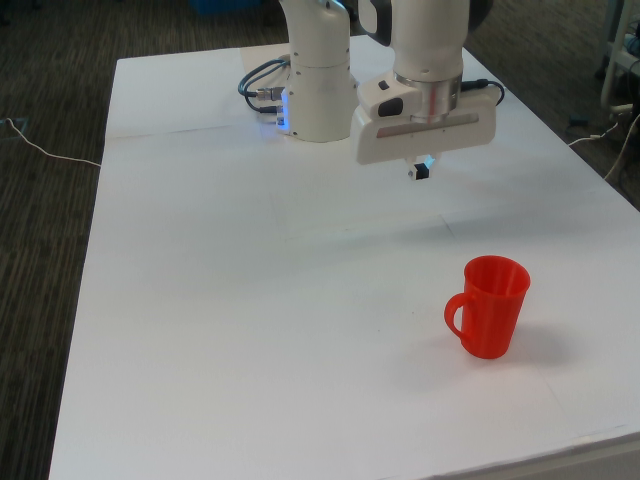} 
        \caption{$\theta_{m3}$}
        \label{fig:mug_real4}
    \end{subfigure}
    \begin{subfigure}{0.16\textwidth}
        \includegraphics[width=\textwidth]{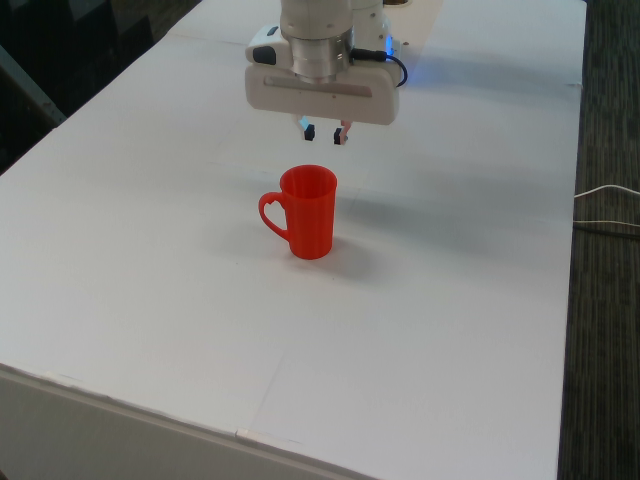}  
        \caption{$\theta_{m4}$}
        \label{fig:mug_real5}
    \end{subfigure}
    \caption{\small Mug lifting task executed with different random camera views in real-world environment. The results for each view are given in Table \ref{tab:mug_success_rate_real}. }
    \label{fig:mug_random_views}
\end{figure}

In order to evaluate the performance of student policies, we tested them under random camera views for cube and mug tasks as shown in Figure \ref{fig:cube_random_views} and Figure \ref{fig:mug_random_views} respectively. 
As $\pi_{s_{t3cam}}$ had demonstrated overall better performance than $\pi_{s_{t2cam}}$ in simulation, we selected it for evaluation in real-world experiments.
Table \ref{tab:cube_success_rate_real} and Table \ref{tab:mug_success_rate_real} show the average success rates of cube and mug lifting tasks for different random views in these figures, respectively.
The success rate is defined as the percentage of successful lifts out of the total number of trials.  

\begin{table}[!h]
    \caption{\small Zero-shot average success rate of cube lifting task for random camera views in Figure \ref{fig:cube_random_views} evaluated in a real-world setup through 20 trials.
     The student policies distilled from the three-camera view and one-camera view teacher policy are denoted as $\pi_{s_{t3cam}}$ and $\pi_{s_{t1cam}}$, respectively.}
    \label{tab:cube_success_rate_real}
    \begin{center}
        \begin{tabular}{p{0.14\textwidth}p{0.12\textwidth}p{0.12\textwidth}}
            \toprule
          Camera View  &  $\pi_{s_{t1cam}}$   &  $\pi_{s_{t3cam}}$      \\
             \midrule
            \hspace{0.5cm}$\theta_{c1}$      & \hspace{0.2cm}40\%     &   \hspace{0.2cm}80\%  \\    
            \hspace{0.5cm}$\theta_{c2}$      & \hspace{0.2cm}60\%     &   \hspace{0.2cm}80\% \\
            \hspace{0.5cm}$\theta_{c3}$      & \hspace{0.2cm}20\%     &   \hspace{0.2cm}60\%  \\
            \hspace{0.5cm}$\theta_{c4}$      & \hspace{0.2cm}35\%     &   \hspace{0.2cm}50\%  \\ 
            \bottomrule
        \end{tabular}
    \end{center}
    \vspace{-15pt}
\end{table}
The results show that the student policy $\pi_{s_{t3cam}}$ outperforms the student policy trained with a single camera view  $\pi_{s_{t1cam}}$ in all camera viewpoints.  
As for the mug lifting task, there is no corresponding $\pi_{s_{t1cam}}$ policy available as the single-camera view teacher was unable to learn the task in the simulation environment. 
Notably, the average success rates were lower than those obtained in the simulation. This can be attributed to the domain gap between simulation and the real-world.
The single-view teacher policy, on the other hand,  was unable to lift the object in any of these random camera views, despite being trained in simulation with the data augmentation method as well.
Moreover, we conducted an analysis to evaluate the impact of the robot state on the performance of the proposed approach. 
The joint states of the robot represented as $q\in \mathbb{R}^{7}$ for its seven degrees of freedom joint angles, were augmented with visual features to train both the teacher and student policies.
The results presented in Table \ref{tab:mug_success_rate_real} demonstrate that incorporating robot joint angles leads to performance improvements for zero-shot policy transfer in a real-world setup.
These findings highlight the need for further research to bridge the gap between simulation and reality and improve the transferability of policies learned in simulation to real-world scenarios.
The video attachment showcases the successful lifting of objects from various random angles and positions, demonstrating the ability of the student policy to handle camera location perturbations in real-time.

\begin{table}[h]
    \caption{\small  Zero-shot average success rate of mug lifting task for random camera views in Figure \ref{fig:mug_random_views} evaluated in a real-world setup through 20 trials. The student policy distilled from the three-camera view teacher policy is denoted as $\pi_{s_{t3cam}}$. The policy distilled from the three-camera view teacher using joint state information is denoted as $\pi_{s_{t3cam+state}}$.}
    
    \label{tab:mug_success_rate_real}
    \begin{center}
        \begin{tabular}{p{0.12\textwidth}p{0.09\textwidth}p{0.09\textwidth}p{0.09\textwidth}}
            \toprule
        Camera View  & $\pi_{s_{t1cam}}$  & $\pi_{s_{t3cam}}$  &  $\pi_{s_{t3cam+state}}$    \\
             \midrule
            \hspace{0.5cm}$\theta_{m1}$  &           \hspace{0.4cm}\---            & \hspace{0.2cm}70\%     &   \hspace{0.2cm}80\%  \\    
            \hspace{0.5cm}$\theta_{m2}$  &           \hspace{0.4cm}\---            & \hspace{0.2cm}70\%     &   \hspace{0.2cm}80\% \\
            \hspace{0.5cm}$\theta_{m3}$  &           \hspace{0.4cm}\---            & \hspace{0.2cm}65\%     &   \hspace{0.2cm}75\%  \\
            \hspace{0.5cm}$\theta_{m4}$  &           \hspace{0.4cm}\---            & \hspace{0.2cm}50\%     &   \hspace{0.2cm}65\%  \\ 

            \bottomrule
        \end{tabular}
    \end{center}
    \vspace{-15pt}
\end{table}

\section{CONCLUSION AND FUTURE WORK}
\label{sec:Conclusion}
In conclusion, we proposed a multi-camera view to single-camera view distillation approach for enhancing the performance of a single-view camera robot policy. 
The proposed method enables a single-view policy to learn from a pre-trained teacher policy, which has been trained with multiple camera viewpoints, ensuring robust feature learning.
To make the student policy robust against unseen visual camera configurations, we utilized a combination of data augmentation techniques and significant variations in camera location and parameters. 
The experimental results demonstrated that our approach can enhance the performance of a single-view camera policy, and successfully lift the challenging object even when the single-camera view teacher policies fail.
Hence, this approach holds promise for increasing the practicality of visual policy learning in scenarios where the use of multiple cameras may not be feasible or where camera calibration is unnecessary.
Future work will focus on investigating the use of point-cloud and depth information to improve performance in zero-shot sim-to-real transfer scenarios.


%


\ifCLASSOPTIONcaptionsoff
  \newpage
\fi



%

\bibliographystyle{IEEEtran}
\bibliography{root}

%




\end{document}